\documentclass[runningheads]{llncs}
\usepackage{graphicx}
\usepackage{amsmath}
\usepackage{amsfonts}
\usepackage{subfigure}

\newtheorem{mydef}{Definition}

\begin{document}
\title{Function Space Pooling For Graph Convolutional Networks}
%
%
\author{Padraig Corcoran}
\authorrunning{P. Corcoran}
%
\institute{School of Computer Science and Informatics, \\ Cardiff University, Wales, UK. \\
\email{corcoranp@cardiff.ac.uk}}
\maketitle              
\begin{abstract}
Convolutional layers in graph neural networks are a fundamental type of layer which output a representation or embedding of each graph vertex. The representation typically encodes information about the vertex in question and its neighbourhood. If one wishes to perform a graph centric task, such as graph classification, this set of vertex representations must be integrated or pooled to form a graph representation.

In this article we propose a novel pooling method which maps a set of vertex representations to a function space representation. This method is distinct from existing pooling methods which perform a mapping to either a vector or sequence space. Experimental graph classification results demonstrate that the proposed method generally outperforms most baseline pooling methods and in some cases achieves best performance.

\end{abstract}
\section{Introduction}
\label{Introduction}
Many real world systems have a relational structure which can be modelled as a graph. These include physical systems where the bodies and joints  correspond to the vertices and edges respectively \cite{sanchez2018graph}; robot swarms where robots and communication links correspond to the vertices and edges respectively \cite{tolstaya2019learning}; and topological maps where locations and paths correspond to the vertices and edges respectively \cite{chen2019behavioral}. Given this, there exists great potential for the application of machine learning to graphs. With the great successes of neural networks and deep learning to the analysis of images and natural language, there has recently been much research considering the application or generalization of neural networks to graphs. In many cases this has resulted in state of the art performance for many tasks \cite{wu2019}.

Graph convolutional is a neural network architecture commonly applied to graphs which consists of a sequence of convolutional layers. The output of a sequence of such layers is a set of vertex representations where each element in this set encodes properties of a corresponding vertex and the vertices in its neighbourhood. In their seminal work, Gilmer et al. \cite{gilmer2017} showed that many different types of convolutional layers can be formulated in terms of a framework containing two steps. In the first step message passing is performed where each vertex receives messages from adjacent vertices regarding their current representation. In the second step, each vertex performs an update of its representation which is a function of its current representation and the messages it received in the previous step. Graph convolution is fundamentally different to the more commonly used image convolution. Unlike an image where each pixel will have an equal number of adjacent pixels (excluding boundary pixels), each vertex in a graph may have a different number of adjacent vertices. Furthermore, unlike an image where the set of pixels adjacent to a given pixel can be ordered, the set of vertices adjacent to a given vertex cannot be easily ordered. Given these facts, generalizing image convolution methods to graphs is non-trivial.

If one wishes to perform a vertex centric task such as vertex classification, then one may operate directly on the set of vertex representations output from a sequence of convolutional layers. However, if one wishes to perform a graph centric task such as graph classification, then the set of vertex representations must somehow be integrated to form a graph representation. We refer to this integration step as \textit{pooling} and it represents the focus of this article. Note that, this step is sometimes referred to as global pooling. Performing pooling represents a challenging problem for a couple of reasons. Firstly, the size of the set of vertex representations will equal the number of vertices in the graph in question and this number will vary from graph to graph. Furthermore, the elements in this set will not be ordered. Therefore the set of vertex representations cannot be directly fed as input to feed-forward or recurrent architecture which require as input an element in a vector space of fixed dimension and an element in a sequence space respectively.

Commonly employed pooling methods include computing summary statistics of the set of vertex representations such as the mean or sum. However these simple pooling methods are not a \textit{complete invariant} in the sense that many different sets of vertex representations may result in the same graph representation leading to weak discrimination power \cite{xu2018}. To overcome this issue and increase discrimination power a number of authors have proposed more sophisticated pooling methods. For example, Ying et al. \cite{ying2018} proposed a pooling method which performs a hierarchical clustering of the set of vertex representations to produce an element in a vector space of fixed dimension.

\begin{figure*}
\begin{center}
\includegraphics[width=12cm]{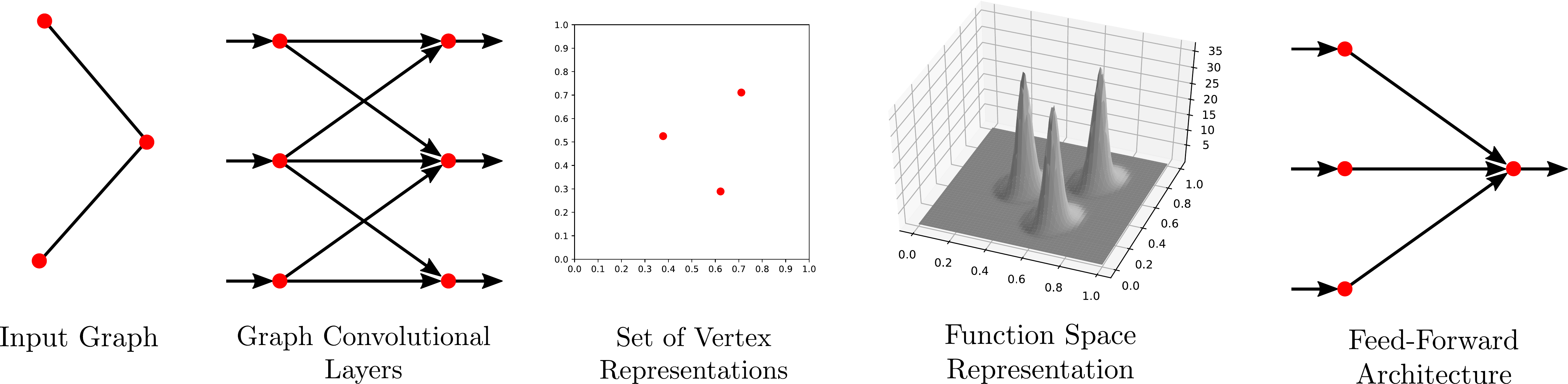}
\caption{The proposed pooling method is illustrated in the context of a complete graph classification architecture. The input graph is first fed to a sequence of graph convolutional layers which outputs a set of vertex representations. The number of elements in this set equals the number of vertices in the original graph. This set is next mapped to a function space representation. This function space representation is then fed to a feed-forward architecture which outputs a predicted graph class.}
\label{fig:classification_arch}
\end{center}
\end{figure*}

In this article we propose a novel pooling method which maps a set of vertex representations to a function space representation. This method is illustrated in Figure \ref{fig:classification_arch} in the context of a complete graph classification architecture. The proposed pooling method is parameterized by a single learnable parameter which controls the discrimination power of the method. This makes the method applicable to both finer and coarser classification tasks which require greater and less discrimination power respectively. The proposed pooling method is inspired by related methods in the field of applied topology which map sets of points in $\mathbb{R}^2$ to function space representations \cite{adams2017}.

The layout of this paper is as follows. Section \ref{sec:related_works} reviews related works on graph convolution architectures and pooling methods. Section \ref{sec:function_space_pooling} describes the proposed pooling method. Section \ref{sec:results} presents an evaluation of this method. Finally section \ref{sec:conclusions} draws some conclusions from this work.

\section{Background \& Related Works}
\label{sec:related_works}
In the following two subsections we review related works on graph convolution architectures and pooling methods.

\subsection{Graph Convolution Architectures}
\label{sec:related_works_graph_conv}
There exist a wide array of graph convolution architectures. In this section we only review those architectures representing theoretical breakthroughs and state of the art. However the interested reader can consult the following review papers for greater details \cite{zhang2018deep,wu2019comprehensive}. Hamilton et al. \cite{hamilton2017} proposed a graph convolution layer known as GraphSAGE which updates a vertex representation by first performing an aggregation of adjacent vertex representations. This aggregation is then concatenated with the current representation of the vertex in question before applying a linear transformation and non-linearity. The authors considered the aggregation functions of mean vertex representation and LSTM (Long Short-Term Memory) applied to a random ordering of vertex presentations. Xu et al. \cite{xu2018} proposed to apply a multi-layer perceptron, as opposed to a single layer which is most common, to the aggregation of adjacent vertex representations and demonstrated that this improve discrimination power. In a later work the same authors \cite{xu2018representation} proposed an architecture known as the jumping knowledge architecture which allows vertices to aggregate information from neighbouring vertices over different ranges. The authors showed that this architecture allows deeper convolutional architectures to be used and outperforms the use of residual connections commonly used in computer vision applications \cite{he2016deep}. 
Given the successes of attention based architectures in natural language processing \cite{vaswani2017attention}, Velickovic et al. \cite{velivckovic2017graph} proposed an attention based architecture for graphs. For a given vertex this architecture allows different weights to be specified for different adjacent vertices.

\subsection{Pooling Methods}
There exist two main categories of pooling methods: those which map the set of vertex representations to a vector space of fixed dimension and those which map the set of vertex representations to a sequence space. The output of these mappings can then be fed as input to a feed-forward or recurrent architecture respectively. We now review pooling methods belonging to each of these categories.

The simplest pooling methods for mapping to a vector space of fixed dimension involve computing summary statistics such as mean and sum of vertex representations \cite{duvenaud2015}. Despite the simple nature of these methods, a recent study by Luzhnica et al. \cite{luzhnica2019graph} demonstrated that in some cases they can outperform more complex methods. To improve discrimination power more sophisticated pooling methods have been proposed. The SortPooling method proposed by Zhang et al. \cite{zhang2018} first sorts the vertices with respect to structural roles in the graph. The vertex representations corresponding to the first $k$ vertices in this order are then concatenated to give a fixed dimensional vector. The value $k$ is a fixed hyper-parameter in the model. Set2Set is a general approach for producing a fixed dimensional vector space representation of a set which is invariant to the order in which the elements are processed \cite{vinyals2015}. Gilmer et al. \cite{gilmer2017} proposed to use this method to perform pooling. Ying et al. \cite{ying2018} proposed a pooling method known as DiffPool which performs a hierarchical clustering of vertex representations and returns an element in a fixed dimensional vector space. Kearnes et al. \cite{kearnes2016} proposed a pooling method based on fuzzy histograms. This method has similarities to that proposed in this article but is formulated in terms of fuzzy theory as opposed to function spaces. The method proposed in this article is in turn distinct. Tarlow et al. \cite{li2015gated} proposed a pooling method which outputs an element in sequence space. Finally, all of the above pooling methods are supervised methods. Many unsupervised pooling methods have also been proposed but we do not review them here \cite{bai2019}.

\section{Function Space Pooling}
\label{sec:function_space_pooling} 

\begin{figure*}
\begin{center}
\subfigure[]{\includegraphics[width=2.75cm]{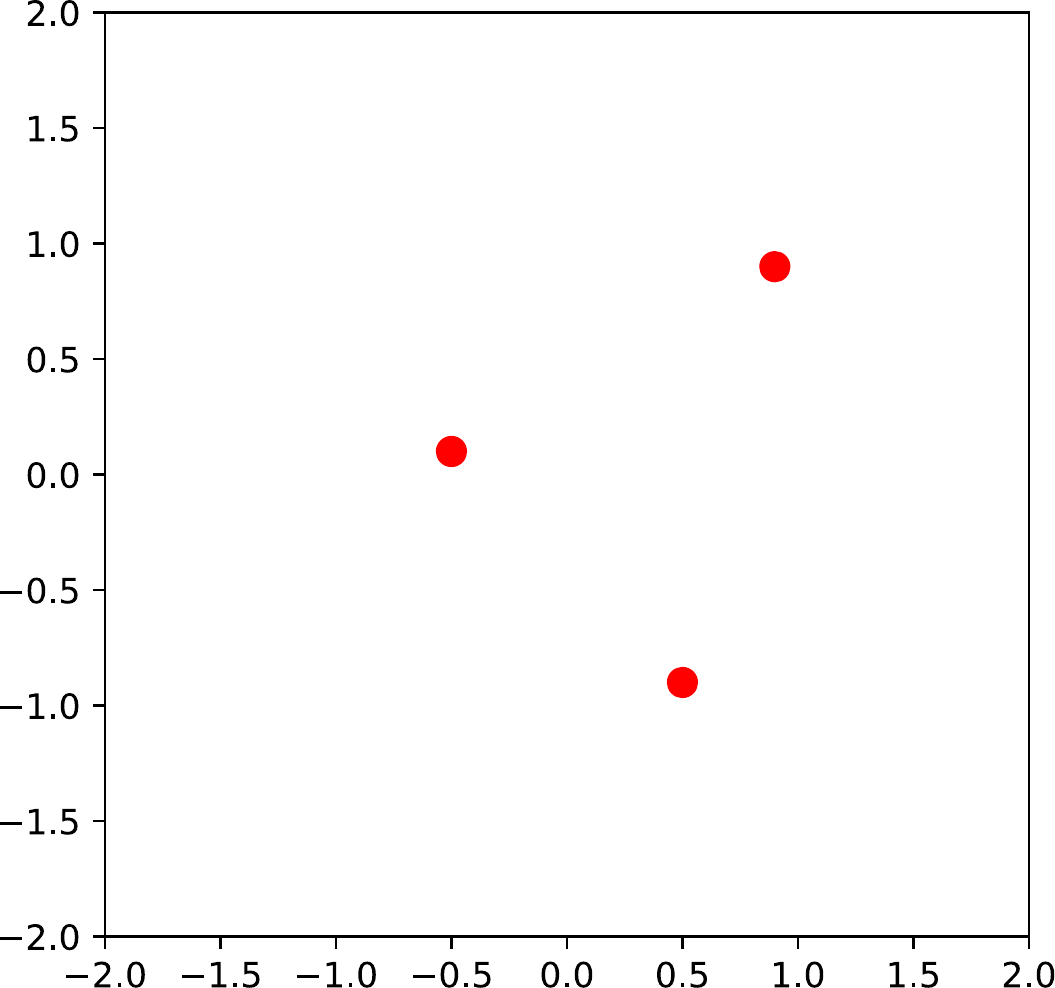}
\label{fig:feature_space_im_1}}
\hspace{0.0cm}
\subfigure[]{\includegraphics[width=2.7cm]{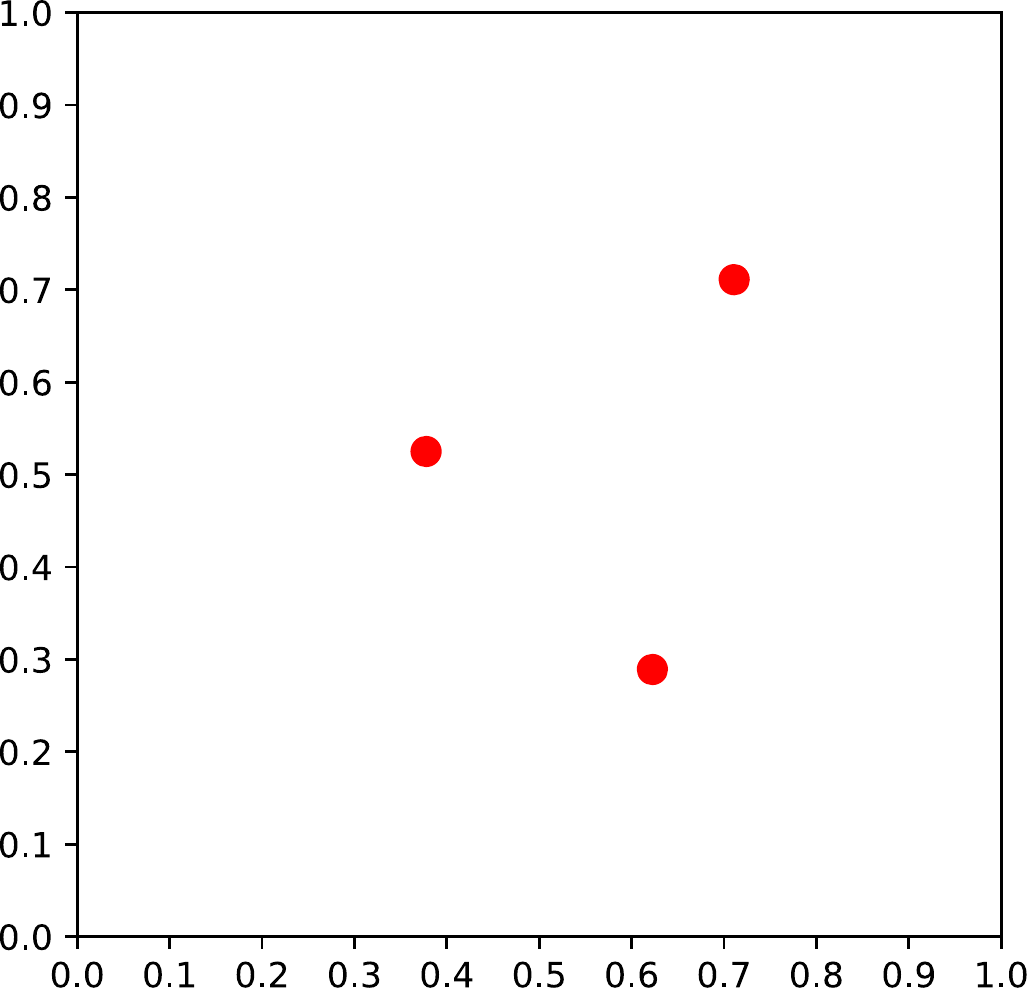}
\label{fig:feature_space_im_2}}
\hspace{0.0cm}
\subfigure[]{\includegraphics[width=2.7cm]{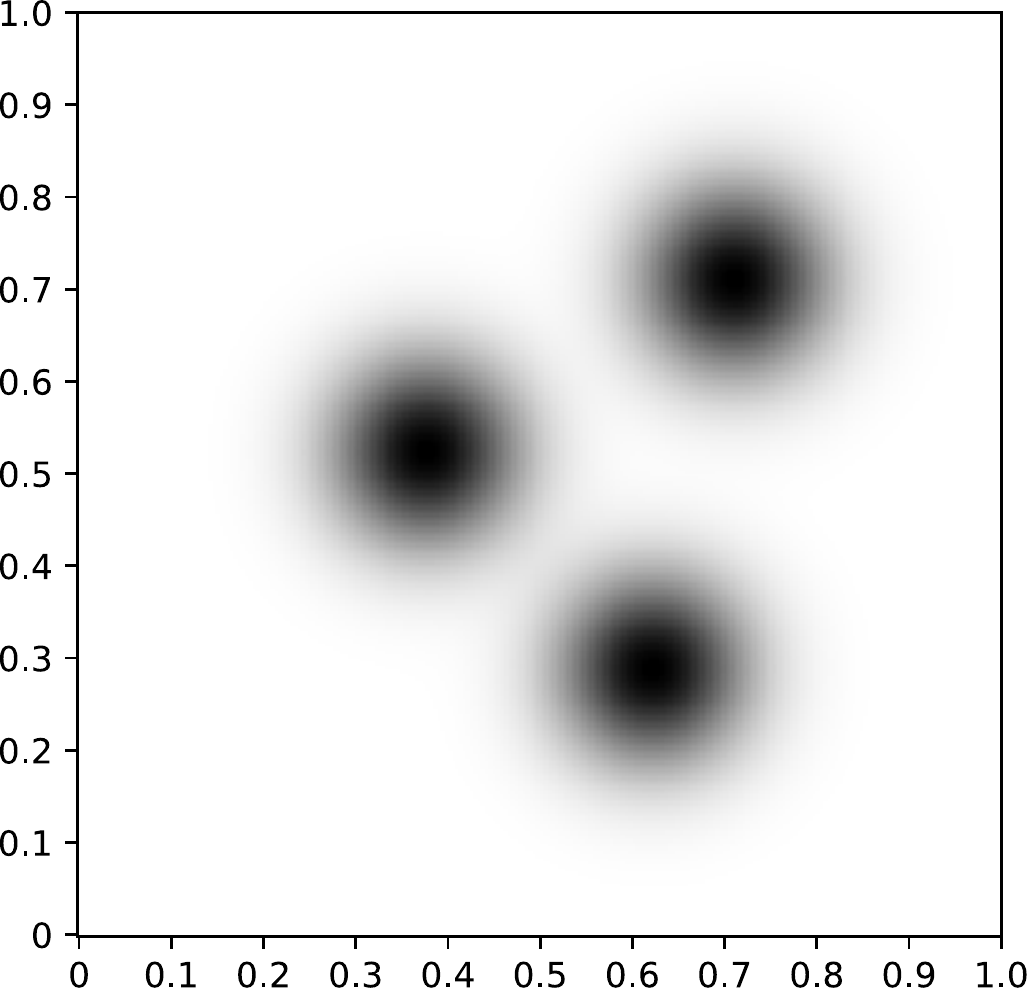}
\label{fig:feature_space_im_3}}
\hspace{0.0cm}
\subfigure[]{\includegraphics[width=2.7cm]{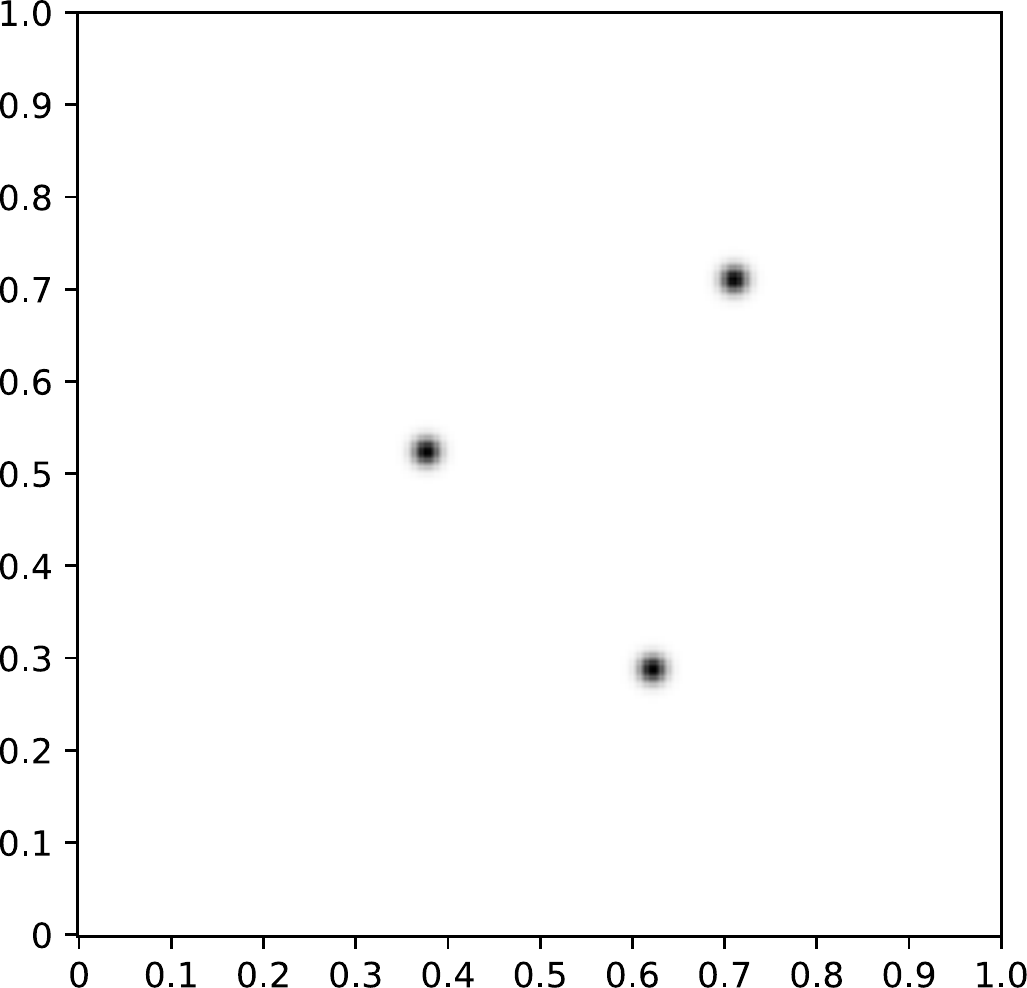}
\label{fig:feature_space_im_4}}
\caption{A set $D$ of vertex representations output from a sequence of convolutional layers is displayed in (a) where each element is represented by a red dot. The result of applying the map $S$ to the set $D$ is the set $s(D)$ displayed in (b). The result of applying the map $F$ to $S(D)$ with the parameter $\sigma=0.005$ is the function $\rho: I \rightarrow \mathbb{R}$ displayed in (c). The result of applying the map $F$ to $S(D)$ with the parameter $\sigma=0.0001$ is the function $\rho: I \rightarrow \mathbb{R}$ displayed in (d).}
\label{fig:evolving_spatial}
\end{center}
\end{figure*}

In this section we present the proposed pooling method. Let graph $G=(V,E)$ denote a graph we wish to classify where $V$ and $E$ are the corresponding sets of vertices and edges respectively. Let $l:V \rightarrow \Sigma$ denote a vertex labelling function that assigns each vertex $v \in V$ a label $l(v)$ in the finite set $\Sigma$.

Let $D$ be the set of vertex representations output from a sequence of convolutional layers applied to $G$. We assume that each element in this set is an element of $\mathbb{R}^n$ where $n$ is a fixed hyper-parameter. The proposed pooling method takes as input $D$ and returns an element in a function space. That is, the method is a map from the space of sets to the space of functions. It contains two steps which we now describe in turn.

The set of vertex representations $D$ is an object in the space of sets which we denote $\Omega$. Let $\textnormal{Sigmoid}: \mathbb{R}^n \rightarrow I$ be the $n$-dimensional Sigmoid function defined in Equation \ref{eq:sigmoid} where $I = \{(x_1, \dots, x_n) \in [0,1]^n \}$ is the $n$-dimensional interval. In the first step of the proposed pooling method we apply the $n$-dimensional Sigmoid elementwise to $D$ to give a map $S: \Omega \rightarrow \Omega$. To illustrate this map consider Figure \ref{fig:feature_space_im_1} which displays an example set $D$ containing three elements in $\mathbb{R}^n$ where $n=2$. The result of applying the map $S$ to this set is illustrated in Figure \ref{fig:feature_space_im_2}.

\begin{equation}
\label{eq:sigmoid}
\textnormal{Sigmoid}(x) = \frac{1}{1 + e^{-x}}
\end{equation}

Let $g_u: \mathbb{R}^n \rightarrow \mathbb{R}$ be a probability distribution. For the purposes of this work we used the $n$-dimensional Gaussian distribution defined in Equation \ref{eq:gaussian} with mean $u$ and variance $\sigma^2$.

\begin{equation}
\label{eq:gaussian}
g_u(x) = \frac{1}{2 \pi \sigma^2} e^{-((x-u)^T(x-u))/2\sigma^2}
\end{equation}

In the second step of the proposed pooling method we apply a map $F:\Omega \rightarrow L^p(I)$ to $S(D)$. Here $L^p(I)$ is the space of real valued functions on $I$ equipped with the $L^p$-norm defined in Equation \ref{eq:lp_norm} \cite{Chr10}. Note that, function addition and subtraction is performed pointwise in this space.

\begin{equation}
\label{eq:lp_norm}
\Vert f \Vert_p = \left( \int_I \vert f(x) \vert^p dx \right)^{1/p}
\end{equation}

The function resulting from the map $F$ is defined in Definition \ref{def:map}. To illustrate this map consider again the example set $S(D)$ illustrated in Figure \ref{fig:feature_space_im_2}. Figure \ref{fig:feature_space_im_3} displays the function $\rho: I \rightarrow \mathbb{R}$ resulting from applying the map $F$ to this set with a $\sigma$ parameter value of $0.005$.

\vspace{.5cm}
\begin{mydef}
\label{def:map}
For $D \in \Omega$ the corresponding function representation $\rho: I \rightarrow \mathbb{R}$ is defined in Equation \ref{eq:function_space}
\begin{equation}
\label{eq:function_space}
\rho(z) = \sum_{u \in S(D)} g_u(z)
\end{equation}
\end{mydef}
\vspace{.5cm}

The elements of $L^p(I)$, and in turn the function representation $\rho: I \rightarrow \mathbb{R}$, are infinite dimensional vector spaces. That is, there are an infinite number of elements in the domain $I$ of $\rho$. We approximate this function as a finite dimensional vector space by discretizing the function domain using a regular grid of elements. For example, the image in Figure \ref{fig:feature_space_im_3} corresponds to a discretizing of the function domain using a $250 \times 250$ grid.

The proposed pooling method is parameterized by $\sigma$ in the probability distribution of Equation \ref{eq:gaussian} where this parameter takes a value in the range $[0, \infty]$. As the value of $\sigma$ approaches $0$ the probability distribution approaches an indicator function on the domain $I$. On the other hand, as the value of $\sigma$ approaches $\infty$ the probability distribution approaches a uniform function on the domain $I$. For example, Figures \ref{fig:feature_space_im_3} and \ref{fig:feature_space_im_4} display the functions $\rho: I \rightarrow \mathbb{R}$ resulting from applying the map $F$ to the set $S(D)$ in Figure \ref{fig:feature_space_im_2} with $\sigma$ parameter values of $0.005$ and $0.0001$ respectively.

The parameter $\sigma$ may be interpreted as follows. As the value of $\sigma$ approaches $0$ the function representation $\rho: I \rightarrow \mathbb{R}$ becomes a sum of indicator functions on the set $D$ of vertex representations. In this case distinct sets $D$ map to distinct functions where the distance between these functions as defined by the norm in Equation \ref{eq:lp_norm} is greater than zero. On the other hand, as $\sigma$ approaches $\infty$, differences between the functions are gradually smoothed out and in turn the distance between the functions gradually reduces. Therefore, one can view the parameter $\sigma$ as controlling the discrimination power of the method. 

\section{Results}
\label{sec:results}
To evaluate the proposed pooling method we considered the task of graph classification on a number of datasets. The layout of this section is as follows. Section \ref{sec:results:architecture} describes the neural network architecture used in all experiments. Section \ref{sec:results:dataset} describes the datasets considered. Section \ref{sec:results:optimization} describes the optimization method used to optimize the network parameters. Finally section \ref{sec:results:classification_accuracy} presents the classification accuracy achieved by the proposed pooling method relative to a number of baseline methods.

\subsection{Network Architecture}
\label{sec:results:architecture}
Recall from section \ref{sec:function_space_pooling} that $G=(V,E)$ denotes a graph we wish to classify and $l:V \rightarrow \Sigma$ denotes a vertex labelling function where $\Sigma$ is a finite set. In order to perform classification of $G$ the following feed-forward neural network architecture was used which consists of six layers.

The first two layers are convolutional layers similarly to the GraphSAGE convolutional layers \cite{hamilton2017}. Only two convolutional layers were used because a number of studies have found that the use of two layers empirically gives best performance \cite{kipf-welling-16}.

Let $\cdot$ denote matrix multiplication and CONCAT denote horizontal matrix concatenation. The $k$th convolutional layer is implemented using Equation \ref{eq:conv_layer} where $A$ is the adjacency matrix corresponding to $G$, $W_k$ are the layer weights and $b_k$ are the layer biases. The weights $W_k$ is a matrix of dimension $2d_{k-1} \times d_k$ where $d_k$ the dimension of the $k$th layer. The biases $b_k$ is a vector of dimension $d_k$. The term $h_0$ denotes a matrix of size $|V| \times |\Sigma|$ where each matrix row equals the one-hot-encoding of an individual vertex label. The term $h_k$ denotes a matrix of size $|V| \times d_k$ where each matrix row equals the representation of an individual vertex output from the $k$th convolutional layer and $d_k$ is the dimension of this representation. Note that, since two convolutional layers are used, $k$ takes values in the set $\lbrace 1,2 \rbrace$. The dimension of the input layer $d_0$ is equal to the number of vertex types since one-hot encoding was used. The dimensions of the two convolutional layers $d_1$ and $d_2$ were both set to $20$.

\begin{equation}
\label{eq:conv_layer}
\begin{split}
h_k \leftarrow & \textnormal{CONCAT}(h_{k-1}, A \cdot h_{k-1}) \\
h_k \leftarrow & \textnormal{ReLu} \left( W_k \cdot h_k + b_k \right)
\end{split}
\end{equation}

The third architecture layer is a fully connected linear layer of dimension $10$. The fourth layer is the pooling method used. The fifth layer is another fully connected linear layer of dimension $20$. The final layer is a softmax function and returns a probability distribution over the classes. The output of the first linear layer equals the input to the pooling method. Therefore the multi-dimensional interval corresponding to the domain of the function $\rho$ in Definition \ref{def:map} is of dimension $10$. We approximate this function as a finite dimensional vector space by discretizing the function domain using a regular grid with $3$ elements in each dimension. This gives a finite dimensional vector space of dimension $3^{10}=59049$.

\subsection{Optimization}
\label{sec:results:optimization}
The model parameters to be optimized in the architecture of section \ref{sec:results:architecture} are the weights and biases of the convolutional layers, the weights and biases of the linear layers and the parameter $\sigma$ of the pooling method. In all experiments the neural network parameters were initialized as follows. All weight matrices in the convolutional and linear layers were initialized using Kaiming initialization \cite{he2015delving}. All biases in the convolutional and linear layers were initialized to zero. Finally, the parameter $\sigma$ in Equation \ref{eq:gaussian} of the pooling layer was initialized to $0.125$.

For loss function Cross Entropy plus an $L^2$ regularization term with weight of $0.2$ was used. The Adam optimization algorithm was used to optimize all model parameters with a learning rate of $1 \times 10^{-3}$ \cite{kingma2014adam}. In all experiments optimization was performed for $350$ data epochs and the model which achieved the minimum loss during this process was returned. In all cases the optimization procedure converged well before $350$ data epochs. 

\begin{table*}[h!]
\centering
\begin{tabular}{c | c | c | c } 
 \textbf{Pooling Method} & \textbf{MUTAG} & \textbf{PROTEINS} & \textbf{ENZYMES}\\ [0.5ex] 
 \hline
 Sum & 0.66 $\pm$ 0.60 & 0.60 $\pm$ 0.18 & 0.26 $\pm$ 0.07 \\ 
 Mean & 0.78 $\pm$ 0.18 & 0.58 $\pm$ 0.16 & 0.30 $\pm$ 0.05 \\
 DiffPool & \textbf{0.85 $\pm$ 0.11} & \textbf{0.73 $\pm$ 0.04} & \textbf{0.32 $\pm$ 0.07} \\
 SortPooling & 0.74 $\pm$ 0.11 & 0.72 $\pm$ 0.05 & 0.23 $\pm$ 0.04 \\
 Set2Set & 0.73 $\pm$ 0.08 & 0.72 $\pm$ 0.04 & 0.30 $\pm$ 0.07 \\
 Function Space & 0.83 $\pm$ 0.11 & \textbf{0.73 $\pm$ 0.19} & \textbf{0.32 $\pm$ 0.06} \\ [1ex] 
 \hline
\end{tabular}
\caption{For each of the MUTAG, PROTEINS and ENZYMES datasets, the mean classification accuracy of 10-fold cross validation for each pooling method are displayed.}
\label{table:mean_accuracy}
\end{table*}

\subsection{Datasets}
\label{sec:results:dataset}
To evaluate the proposed pooling method we used three graph classification datasets. The datasets in question are commonly used to evaluate graph classification methods and were obtained from the TU Dortmund University graph dataset repository \cite{KKMMN2016}.

The first dataset was the MUTAG dataset which consists of 188 graphs corresponding to chemical compounds where there are $7$ distinct types of vertices. The classification problem is binary and concerns predicting if a chemical compound has mutagenicity or not \cite{debnath1991}.

The second dataset was the PROTEINS dataset which consists of $1113$ graphs corresponding to protein molecules where there are $3$ distinct types of vertices. The classification task is binary and concerns predicting if a protein is an enzyme or not \cite{borgwardt2005}.

The third dataset was the ENZYMES dataset which consists of $600$ graphs corresponding to enzymes where there are $3$ distinct types of vertices. The classification task is multi-class and concerns predicting enzyme class where there are $6$ distinct classes.

\subsection{Classification Accuracy}
\label{sec:results:classification_accuracy}
The proposed pooling method was benchmarked against the following five baseline pooling methods: mean vertex representation, sum of vertex representations, DiffPool by Ying et al. \cite{ying2018}, SortPooling by Zhang et al. \cite{zhang2018} and Set2Set by Vinyals et al. \cite{vinyals2015}. As described in the background and related works section of this paper, these are some of the most commonly used pooling methods.

For all baseline models we used a neural network architecture similar to that described in section \ref{sec:results:architecture} with the exception that the pooling layer was replaced and the dimension of the linear layer before this layer was changed from $10$ to $20$. All experiments were implemented in Python3 using the PyTorch library \cite{paszke2017automatic} and run on an Nvidia GeForce RTX 2080 GPU. For the baseline pooling methods we used the corresponding implementations available in the \textit{PyTorch Geometric} Python library \cite{Fey/Lenssen/2019}. 

For each dataset considered we computed the mean accuracy of 10-fold cross validation for each pooling method. The results of this analysis are displayed in Table \ref{table:mean_accuracy}. For each dataset, the proposed pooling method outperformed most baseline methods and achieved equal best performance on two of the three datasets. This demonstrates the utility of the proposed pooling method.

\section{Conclusions}
\label{sec:conclusions}
Pooling is a fundamental type of layer in graph neural networks which involves compute a representation of the set of vertex representations output from a sequence of convolutional layers. In this work we proposed a novel pooling method which computes a function space representation of the set of vertex representations. This method is distinct from existing pooling methods which compute either a vector or sequence space representation.

Experimental results on a number of graph classification benchmark datasets demonstrate that the proposed method generally outperforms most baseline pooling methods and in some cases achieves best performance. The benchmark datasets in question contain graphs corresponding to molecules and chemical compounds which are the most common types of dataset used to evaluate graph classification methods. Despite this fact, the proposed pooling method is general in nature and can be applied to any type of graph. Finally, the authors hope this work will serve as a platform for future work investigating the use of function space representations for pooling.
%
%
%
%

\end{document}